\documentclass{article}

\PassOptionsToPackage{nonatbib, preprint}{neurips_2025}
\usepackage{neurips_2025}
\usepackage{multirow}
\usepackage{booktabs}
\usepackage{graphicx}
\usepackage[utf8]{inputenc} 
\usepackage[T1]{fontenc}    
\usepackage{hyperref}       
\usepackage{url}            
\usepackage{subfig}
\usepackage{amsfonts}       
\usepackage{nicefrac}       
\usepackage{microtype}      
\usepackage{xcolor}         
\usepackage{colortbl}
\usepackage{utfsym}
\usepackage{bbding}
\usepackage{booktabs}       
\usepackage[backend=biber, style=numeric]{biblatex}
\usepackage{xcolor}
\usepackage{graphicx}
\usepackage{wrapfig}

\usepackage[most]{tcolorbox}
\usepackage{array}
\usepackage{caption}
\usepackage{enumitem}

\definecolor{deeporange}{rgb}{0.8, 0.3, 0.0}

\usepackage{tikz}
\usepackage{xcolor}
\usepackage{hyperref}
\usepackage{enumitem}

\definecolor{light-gray}{gray}{0.6}
\definecolor{front-color}{HTML}{F5FFFA}
\definecolor{Gray}{gray}{0.93}

\title{CVBench: Benchmarking Cross-Video Synergies for Complex Multimodal Reasoning}
\author{
  Nannan Zhu$^{1}$$^{\dagger}$,\ 
  Yonghao Dong$^{1}$$^{\dagger}$,\ 
  Teng Wang$^{2}$$^{*}$,\ 
  Xueqian Li$^{1}$,\ 
  Shengjun Deng$^{3}$,\ 
  Yijia Wang$^{1}$ \\
  \textbf{Zheng Hong}$^{1}$,\ 
  \textbf{Tiantian Geng}$^{4}$,\ 
  \textbf{Guo Niu}$^{3}$,\ 
  \textbf{Hanyan Huang}$^{1}$,\ 
  \textbf{Xiongfei Yao}$^{3}$,\ 
  \textbf{Shuaiwei Jiao}$^{3}$\\
  $^{1}$ Sun Yat-sen University \quad
  $^{2}$ University of Hong Kong \quad \\
  $^{3}$ Foshan University  \quad
  $^{4}$ University of Birmingham \\
}

\begin{document}

\maketitle

\begin{abstract}
While multimodal large language models (MLLMs) exhibit strong performance on single-video tasks (e.g., video question answering), their capability for spatiotemporal pattern reasoning across multiple videos remains a critical gap in pattern recognition research. However, this capability is essential for real-world applications, including multi-camera surveillance and cross-video procedural learning. To bridge this gap, we present CVBench, the first diagnostic benchmark designed to assess cross-video relational reasoning rigorously. CVBench comprises 1,000 question-answer pairs spanning three hierarchical tiers: \textit{cross-video object association} (identifying shared entities), \textit{cross-video event association} (linking temporal or causal event chains), and \textit{cross-video complex reasoning} (integrating commonsense and domain knowledge). Built from five domain-diverse video clusters (e.g., sports, life records), the benchmark challenges models to analyze and integrate spatiotemporal patterns from dynamic visual streams. Extensive evaluation of 10+ leading MLLMs (including GPT-4o, Gemini-2.0-flash, Qwen2.5-VL) under zero-shot or chain-of-thought prompting paradigms. Key findings reveal stark performance gaps: even top models, such as GPT-4o, achieve only 63.5\%  accuracy on causal reasoning tasks, compared to the 91.3\% accuracy of human performance. Crucially, our analysis reveals fundamental bottlenecks inherent in current MLLMs architectures, notably deficient inter-video context retention  and poor disambiguation of overlapping entities. CVBench establishes a rigorous framework for advancing pattern recognition methodologies in multi-video scenarios, providing architectural insights for next-generation models. The data and evaluation code are available at: \url{https://github.com/Hokhim2/CVBench}.
\end{abstract}

\section{Introduction}

The ability to reason across multiple video streams is fundamental to real-world applications, such as multi-camera surveillance systems, distributed procedural learning, and multi-view activity analysis~\cite{Wu2025}. While multimodal large language models (MLLMs) have demonstrated significant progress in single-video understanding~\cite{CAI2025111670}, their capacity for spatiotemporal pattern reasoning across videos remains a critical and underexplored challenge in pattern recognition research~\cite{ZHENG2025111319}. This capability requires models to identify and analyze relational patterns across distinct spatiotemporal contexts~\cite{11008449}, such as entity correspondences, event sequences, and causal links. The research gap stems not merely from a lack of single-video benchmarks, but from the inherent limitations of existing evaluations in capturing cross-video relational dynamics, which are central to real-world pattern recognition systems.

Current MLLMs face fundamental barriers in cross-video scenarios:
1) \textit{Inter-video context linking}: models lack mechanisms to persistently integrate and retain object/event states cross-video separated by time and space~\cite{Li10658165,11093290}; 2) \textit{Entity disambiguation}: overlapping or visually similar entities cross-video lack unique spatiotemporal grounding, causing identification errors; 3) \textit{Temporal-causal modeling}: existing architectures poorly capture long-range dependencies and causal links spanning asynchronous video sequences~\cite{11095110}.

\begin{figure*}[t]
    \centering
   \includegraphics[width=0.95\linewidth]{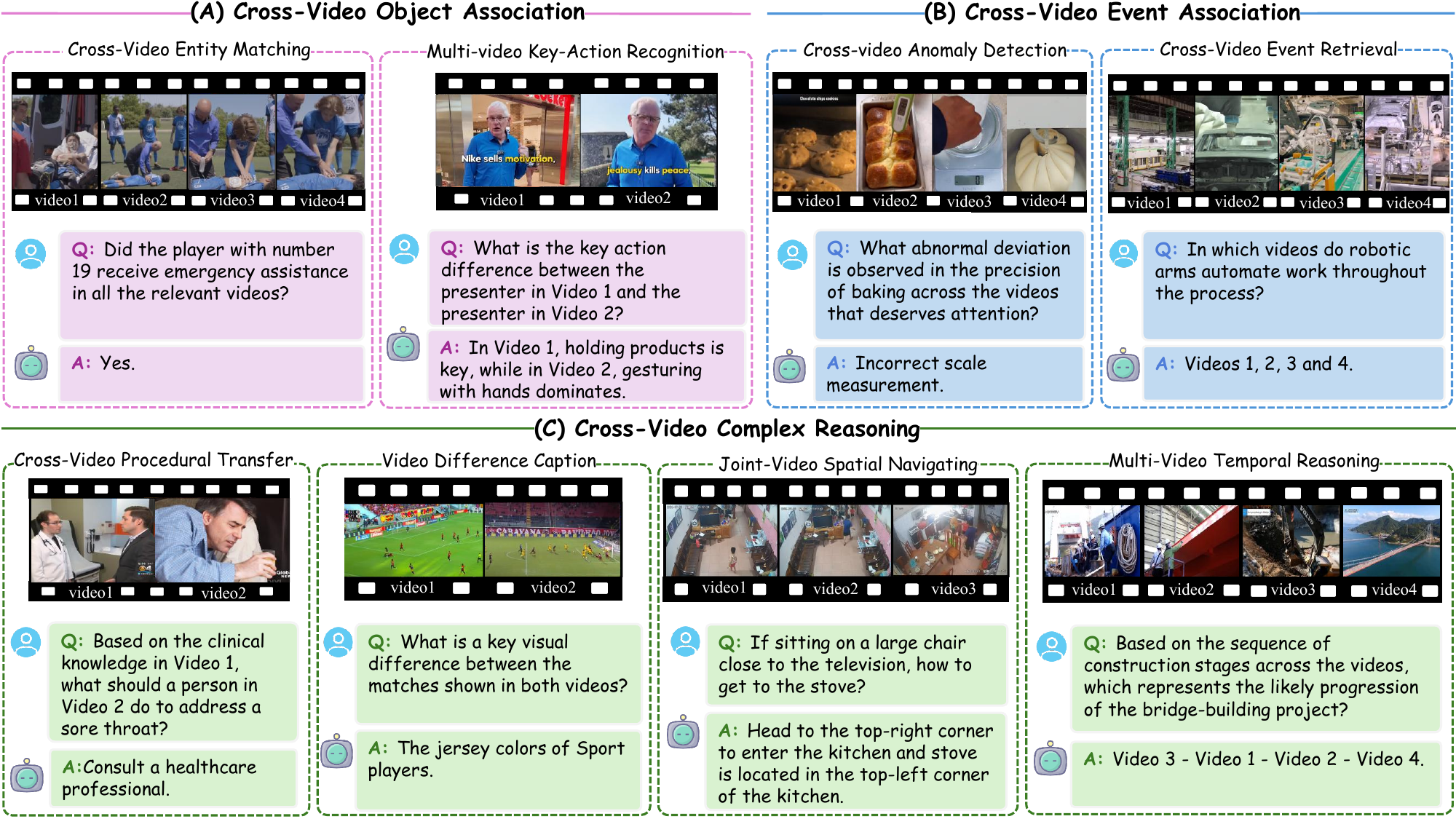} 
    \caption{Illustration of CVBench task definiton. CVBench focuses on three core tasks that are crucial for real-world applications: (A) cross-video object association (identifying shared entities across different videos), (B) cross-video event association (linking temporal or causal event chains that span across video sequences), and (C) cross-video complex reasoning (integrating commonsense to understand and reason about complex interrelationships between video content). The benchmark format is multi-choice QA, and here we only show the correct answer for visualization.}
    \label{abstract}
    \vspace{-1em}
\end{figure*}

To address these limitations and provide a diagnostic framework for the community, we introduce CVBench, the first comprehensive benchmark designed specifically for evaluating cross-video relational pattern recognition. As illustrated in Fig.~\ref{abstract}, CVBench structures the evaluation around three hierarchically organized tasks:\textit{cross-video object association} for verifying entity persistence across varying visual contexts, 
\textit{cross-video event association} for modeling temporal sequences and causal relationships cross-video, and
\textit{cross-video complex reasoning} for integrating visual cues and external knowledge for hierarchical spatiotemporal understanding.
CVBench comprises 1,000 QA pairs sampled from five diverse domains (artistic performances, sports competitions,  films and television, life records,  and knowledge) to ensure generalization. This domain diversity challenges models to reason across dynamic contexts, requiring {spatio-temporal integration} and {multi-view synthesis}. Expert annotations capture real-world complexities by modeling entities, events, and inter-video relationships. This rigorous foundation enables diagnostic evaluation of cross-video reasoning capabilities.

We evaluate over ten state-of-the-art closed-sourced and open-sourced MLLMs, including GPT-4o, Gemini series, and Qwen2.5-VL, under zero-shot and chain-of-thought prompting paradigms. Our evaluation reveals significant performance gaps in these models. Specifically, GPT-4o achieves only 60.2\% accuracy on causal reasoning tasks, which is lower than the human-level performance of 91.3\%. Additionally, these models struggle with inter-video context retention and difficulties in disambiguating overlapping entities, which hinders their ability to integrate information from multiple sources. These findings underscore the limitations of current MLLMs in handling complex cross-video tasks, highlighting the need for more advanced architectures to address challenges like long-term context retention and entity disambiguation. 

The contributions of this paper can be summarized as:

\begin{itemize}
    \item 

We introduce CVBench, the first diagnostic benchmark explicitly targeting cross-video understanding through object association, event association, and complex reasoning.
 \item 

We release a carefully curated multi-domain dataset with comprehensive annotations, enabling rigorous stress-testing of spatiotemporal integration and multi-view synthesis capabilities.
\item 
{We provide a systematic analysis of state-of-the-art MLLMs on CVBench, exposing critical performance gaps and identifying key research directions for advancing cross-video pattern reasoning in the field.
}

\end{itemize}

\begin{figure*}[t!]
    \centering
    \includegraphics[width=0.95\linewidth]{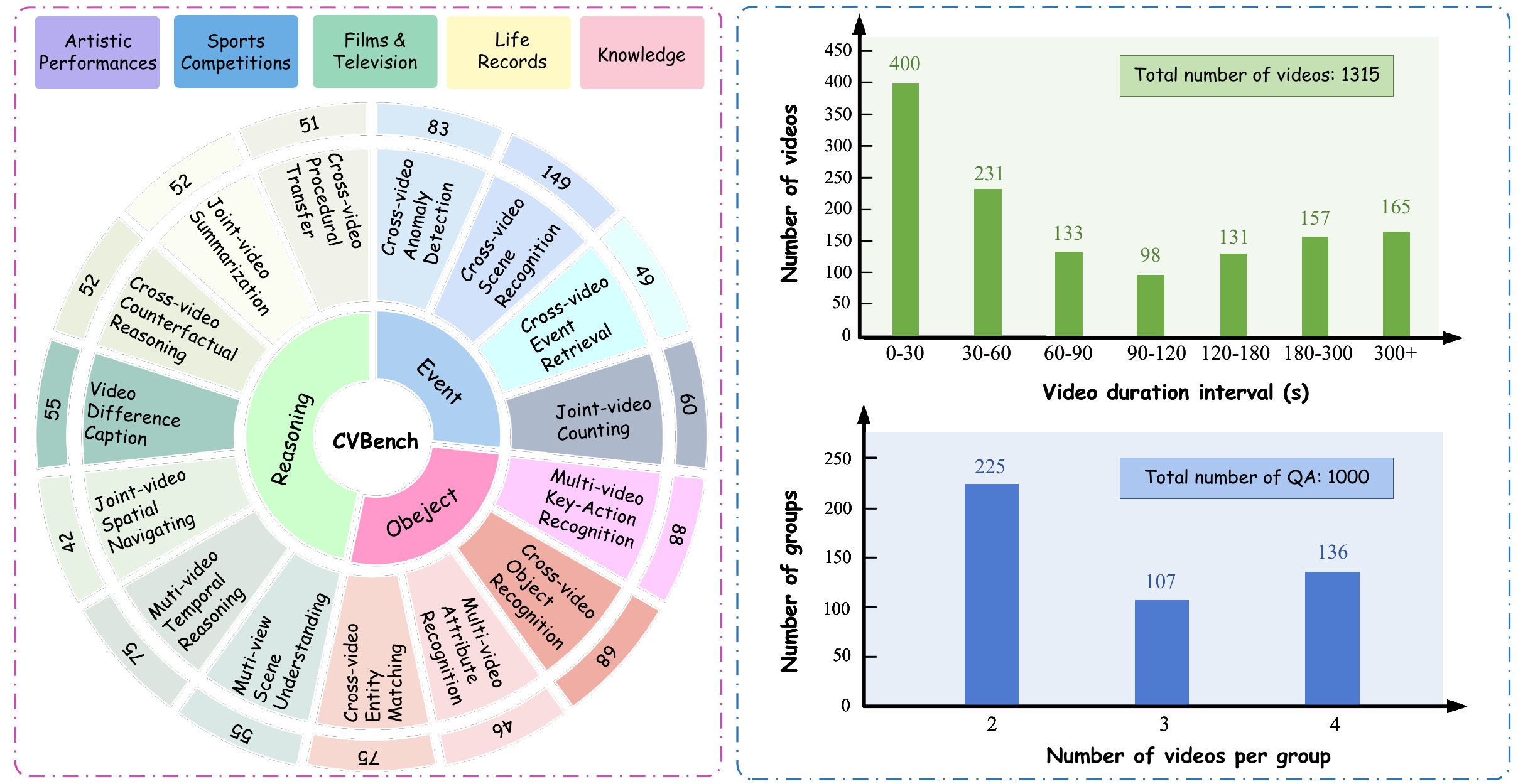}
    \caption{CVBench statistics and task categories. CVBench comprises three primary task categories and 15 video sub-categories across diverse video domains. We present the distribution of video durations, the number of videos, and representative domains.}
    \label{fig:categories}
\end{figure*}
\section{Related Work}

\subsection{Multimodal Large Language Models}
The evolution of MLLMs has progressed from image-based~\cite{Zhang10547418} to video-capable architectures~\cite{Fu11093290, ZHANG2026111925}, marked by two main improvements: 1) Long-range context modeling that enables efficient handling excessively long temporal contexts~\cite{Yue11093104, Song10657734}; 2) Spatio-temporal fusion that integrates both appearance and motion dynamics with language reasoning~\cite{Ren10656135, Zhang11094660, Liu11094049}. Despite these advances, current MLLMs remain architecturally constrained to single-video scenario, lacking mechanisms for cross-video interactions, exhibiting inadequate spatiotemporal grounding for overlapping entities, and demonstrating limited modeling of temporal event chains. These fundamental limitations restrict their applicability to multi-camera surveillance and cross-video applications~\cite{Yang11037495}.

\subsection{Video Understanding Benchmarks}
A comprehensive and objective benchmark is essential for evaluating MLLMs, enabling the comparison and investigation of the performance of various models~\cite{SHAO2026112724}. Recent advances in multimodal learning have led to the development of several benchmarks for video understanding tasks~\cite{ZHONG2025111035}. However, existing video evaluation frameworks exhibit critical limitations in assessing cross-video reasoning capabilities. First, \textit{general multimodal frameworks} are predominantly with broad domain coverage, such as Video-MMMU and Video-MME~\cite{Fu11093290}), to evaluate instruction-following capabilities but lacking cross-video relational tasks. Second, \textit{long-term video benchmarks} contain longer videos with multiple sub-events,  however, are confined to single source or similar creators, such as VideoTree~\cite{Wang11094563}, Video-Bench~\cite{Ning11284911}, and Chapter-Llama~\cite{Ventura11092338}. These deficiencies fundamentally limit their applicability to real-world scenarios in multi-camera surveillance and cross-video procedural learning. 

To address these gaps, we introduce CVBench, the first diagnostic framework for cross-video relational reasoning across different sources. CVBench pioneers a three-tiered evaluation of spatiotemporal integration capabilities, encompassing object-level association, event-level association and complex reasoning.

\begin{figure*}[t!]
    \centering
    \includegraphics[width=0.97\linewidth]{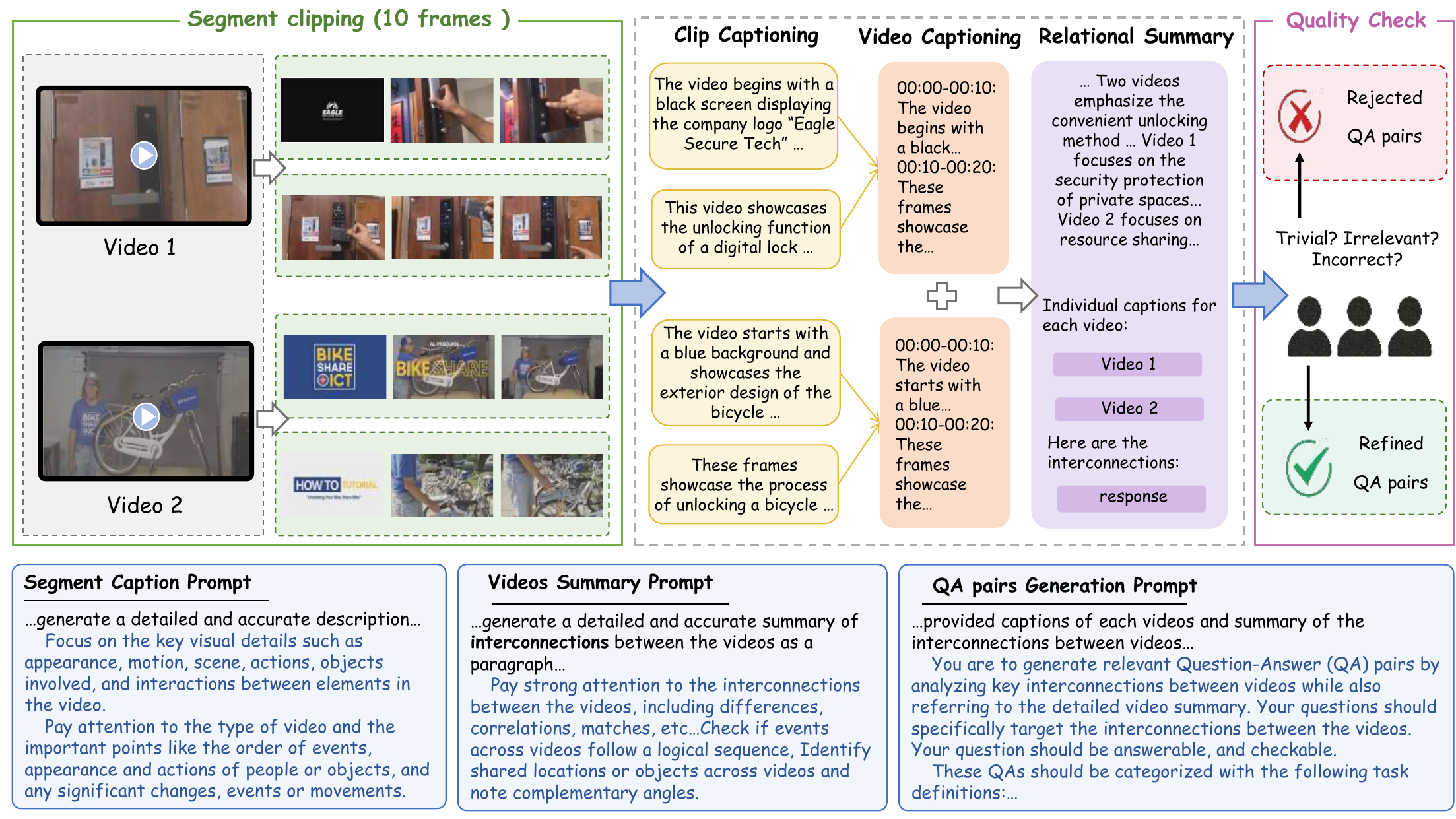}
    \caption{The pipeline for question-answer (QA) pairs annotations in CVBench. The QA generation pipeline is streamlined as four steps: video segment captioning, relational summary annotation, question-answer generation and quality control. } 
    \label{fig:qapair}
\end{figure*}

\section{CVBench: Cross-Video Reasoning Benchmark}

\subsection{Task Definition}

\label{subsec:task_definition}
CVBench pioneers the evaluation of cross-video relational reasoning through a three-tiered diagnostic framework, encompassing \textit{object association}, \textit{event association}, and \textit{complex reasoning}. This taxonomy systematically addresses critical gaps in real-world video understanding by requiring models to establish inter-video relationships through spatiotemporal pattern recognition beyond single-video analysis. The framework comprises 15 subtasks (detailed in Appendix) categorized as follows:

\begin{itemize}
    \item \textbf{Object association}: Cross-video object recognition, multi-video attribute recognition, joint-video counting, and cross-video entity matching. These require consistent identification of entities across domains or viewpoints despite occlusion or deformation.
    
    \item \textbf{Event association}: Cross-video anomaly detection, scene recognition, key-action recognition, event retrieval, multi-view scene understanding, and temporal reasoning. These demand fusion of spatiotemporal signatures across videos to reconstruct event dynamics.
    
    \item \textbf{Complex reasoning}: Joint-video spatial navigating, video difference captioning , cross-video counterfactual reasoning, joint-video summarization, and procedural transfer. These necessitate higher-order synthesis including spatial imagination, fine-grained reasoning, causal inference and adaptive knowledge transfer.
\end{itemize}

\subsection{Dataset Collection}

CVBench is constructed with three-step process: video curation, question-answer annotation, and quality control. First, diverse videos are selected to ensure broad scenario coverage. Next, question-answer pairs are crafted to capture complex reasoning and cross-video dependencies. Finally, rigorous quality control removes ambiguous or trivial entries, ensuring the dataset’s relevance and reliability for evaluation.

\paragraph{Video curation.}
We employ task-driven sampling from YouTube to construct diverse video groups (at most four videos per group). Detailed statistic are shown in Fig.~\ref{fig:categories}. We ensure source diversity through the inclusion of multi-creator channels (videos with similar or same creators), cross-domain associations (videos across different domain but with shared topics or intrinsic relations), and the segmentation of long videos (sequential occurrence of event-level activities). Specifically, we measure video quality under the following four curation criteria: 

\begin{itemize}
    \item Multi-domain coverage, including daily life, sports, surveillance and other commen scenrios;
    \item Intrinsic inter-video relationships, such as differences, similarities, co-occurrence, causality causal, temporal order;
    \item Duration constraints, where videos are primarily within 10 minutes in length, and those exceeding 10 minutes account for less than 5\% of the dataset;  
    \item Visual quality and availability, requiring videos to have a resolution of at least 720p and a frame rate of 24 fps.
\end{itemize}

\begin{table*}[t!]
\centering
\small
\resizebox{\linewidth}{!}{
\begin{tabular}{lccccccc}
\toprule
\textbf{Benchmark}        & \#\textbf{Video}   & \#\textbf{QA}  & \#\textbf{Video Per QA} & \textbf{Duration (s)}   & \textbf{Multi-level} &   \textbf{CV.U} & \textbf{CV.R}    \\ \midrule
{TempCompass \cite{tempcompass2024}}          &   410          & 500        & 1   &   11.4                & \textcolor{red}{\XSolidBrush}  & \textcolor{red}{\XSolidBrush}  & \textcolor{red}{\XSolidBrush}  \\
{MSVD-QA \cite{xu2017video}}          & 504             & 13,157        & 1   &   9.8                 & \textcolor{red}{\XSolidBrush}  & \textcolor{red}{\XSolidBrush}  & \textcolor{red}{\XSolidBrush}  \\
{Video-MME \cite{fu2024video}}        & 900             & 2,700       & 1   &     1017.9                      & \textcolor{green}{\checkmark}   & \textcolor{red}{\XSolidBrush}  & \textcolor{red}{\XSolidBrush}  \\
{NExT-QA \cite{xiao2021next}}          & 1,000            & 8,564      &  1  &    39.5                      & \textcolor{red}{\XSolidBrush}   & \textcolor{red}{\XSolidBrush}  & \textcolor{red}{\XSolidBrush}  \\
{AutoEval-Video \cite{chen2024autoeval}}    & 327             & 327     & 1   &       14.6                       & \textcolor{red}{\XSolidBrush}  & \textcolor{red}{\XSolidBrush}  & \textcolor{red}{\XSolidBrush}  \\
EgoSchema \cite{egoschema2023} & 5,063 & 5,063 & 1 & 180.0 & \textcolor{red}{\XSolidBrush} & \textcolor{red}{\XSolidBrush} & \textcolor{red}{\XSolidBrush} \\
ActivityNet-QA \cite{ActivityNet2019}    & 800             & 8,000     & 1   &       111.4                       & \textcolor{red}{\XSolidBrush}  & \textcolor{red}{\XSolidBrush}  & \textcolor{red}{\XSolidBrush}  \\
\rowcolor{gray!20}
\textbf{CVBench (Ours)}          & 1,315            & 1,000          & 2 $\sim$ 4   &    106.6       & \textcolor{green}{\checkmark} & \textcolor{green}{\checkmark} & \textcolor{green}{\checkmark}  \\ \bottomrule
\end{tabular}
}
\caption{Comparison of video understanding benchmarks. CV.U and CV.R denote cross-video understanding and cross-video reasoning, respectively. ``Multi-level” indicates videos with multiple duration levels.}

\small
\label{comwithc}
\end{table*}

\subsubsection{QA generation}

\paragraph{QA generation.}

The multi-choice QA generation pipeline comprises four sequential stages (see Fig.~\ref{fig:qapair}). First, videos are segmented using an adaptive frame sampling strategy, where the sampling interval increases with video length to optimize both semantic coverage and annotation efficiency. Next, detailed segment-level captions are generated, emphasizing key visual elements such as appearance, actions, and interactions. These captions are then aggregated into comprehensive video-level descriptions, followed by a relational summary that explicitly captures interconnections across videos, including entity state transitions, causal relationships, and contextual complementarities or contradictions. Leveraging these multi-level annotations, we synthesize at least two multi-choice QA (with one correct choice and three disturbance) pairs per group, each designed to require cross-video dependencies, incorporate multiple adversarial distractors (entity, temporal, or causal), and involve multi-hop reasoning. These constraints are enforced through prompt engineering with GPT-4o to ensure the QA pairs are non-trivial and task-aligned. Finally, all QA pairs undergo rigorous quality control, where trivial, irrelevant, or incorrect pairs are filtered out, resulting in a challenging and reliable QA dataset. The prompts used for caption generation, relational summary construction, and QA pair synthesis with GPT-4o are illustrated in Fig.~\ref{fig:qapair}.

\paragraph{Quality control.}

In terms of video collection, we hire eight annotators with bachelor or master degrees follow a rigorous three-stage collection protocol. First, an initial keyword-based retrieval process yields 1,500 candidate videos. Next, a gap-filling stage augments underrepresented domains to enhance coverage. Finally, a quality-controlled replacement procedure is activated when QA pair quality are not met, ensuring the reliability of the dataset. The resulting curated dataset comprises 1,315 videos with a stratified duration distribution: short, mid-short, mid, and long videos constitute 30.4\%, 17.5\%, 27.5\%, and 24.4\% samples, respectively. 

For the quality control of QA pairs, we follow a two-stage human checking protocol. First, eight annotators verify QA pairs against a set of 15 task-specific criteria, initiating corrective actions as needed: manual revision for low-quality questions, option correction for unclear disturbance term, recollect videos for topic inconsistencies, etc. Then, five new annotators conduct a comprehensive re-examination (200 QA pairs per annotator), verifying question compliance, answer correctness, and the absence of hallucinations.

\subsection{Comparison with Previous Benchmarks}

CVBench differs significantly from traditional single-video and multi-image benchmarks, as shown in Table~\ref{comwithc}. While single-video benchmarks focus on isolated content, they fail to evaluate models' ability to integrate information across multiple videos and understand temporal and spatial relationships. Multi-image benchmarks, although involving multiple inputs, do not address the dynamic, temporal aspects of videos or the complex reasoning required for cross-video understanding.
In contrast, CVBench is explicitly designed to assess cross-video relationships, incorporating tasks that require models to reason across multiple video streams. With diverse video domains and task types, CVBench provides a comprehensive evaluation framework that advances cross-video understanding and reasoning. Unlike existing benchmarks, CVBench evaluates cross-video understanding (CV.U) and reasoning (CV.R), ensuring a more holistic assessment of multimodal models' capabilities in real-world applications.

\begin{table*}[t!]
\centering
\small
    \renewcommand\arraystretch{1.0}
    \setlength{\tabcolsep}{2.3 mm}
\begin{tabular}{lcccccccc}
\toprule
Categories&     
  Average &
  M. SU &
  M. TR &
  J. SN &
  VDC &
  C. CR &
  J. S &
  C. PT \\ 
(QA pair numbers) &
  (382)&
  (55) &
  (75) &
  (42) &
  (55)&
  (52) &
  (52) &
  (51)\\  
\midrule
  \multicolumn{1}{l}{Human} 
&91.3
&80.0
&100.0
&83.3
&100.0
&100.0
&100.0
&83.3
\\

\multicolumn{1}{l}{Random Choice} 
   &26.2
   &21.8
   &30.7
   &21.4
   &21.8
   &30.8
   &25.0
   &29.4
   \\
\midrule
    \rowcolor{gray!20}
\multicolumn{9}{c}{\textbf{\textit{Closed-source MLLMs}}} \\
\multicolumn{1}{l}{GPT-4o-mini} 
&64.4
& 87.3 	
& 41.3	
& 47.6	
& 58.2	
&67.3	
& 80.8	
&70.6
   \\
\multicolumn{1}{l}{GPT-4o} 
&69.1
& 89.1	
& 58.7	
& 61.9	
& 61.8	
&63.5	
& 82.7	
&66.7
   \\
\multicolumn{1}{l}{Gemini-1.5-flash}   
&62.8
& 85.5 	
&49.3	
&	38.1
& 61.8	
& 57.7	
&80.8	
& 64.7
        \\ 
\multicolumn{1}{l}{Gemini-2.0-flash} 
&69.4
& 89.1	
&58.7	
& 50.0	
& 70.9	
&67.3	
&76.9	
& 66.7
      \\ 
      \midrule
       \rowcolor{gray!20}
\multicolumn{9}{c}{\textbf{\textit{Open-source MLLMs}}} \\ 
Qwen2.5-Omni-7B \cite{xu2025qwen25omnitechnicalreport} 
&52.4
&
  70.9 	&
  22.7 	&
  31.0 &	
  56.4 	&
  57.7 	&
  75.0 	&
  60.8 \\ 
Qwen2.5-VL-7B \cite{bai2025qwen25vltechnicalreport} 
&51.3
&
  80.0 &
  22.7 &
  26.2 &
  50.9 &
  55.8 &
  69.2 &
  60.8 \\ 
LLaVA-OneVision-7B \cite{li2024llama} 
&52.6
&
  83.6 &
  40.0 &
  38.1 &
  45.5 &
  42.3 &
  61.5 &
  52.9 \\ 
{\color{black}VideoLLaMA3-7B \cite{cheng2024videollama}}
&57.6
&
  78.2 &
  29.3 &
  35.7 &
  60.0 &
  59.6 &
  80.8 &
  64.7 \\ 
{\color{black}InternVL2.5-8B \cite{chen2024internvl}}
&59.4
&
  83.6 &
  26.7 &
  50.0 &
  60.0 &
  69.2 &
  67.3 &
  68.6 \\ 
Phi-4-Multimodal-5B \cite{abouelenin2025phi} 
&49.7
&
  69.1 &
  26.7 &
  31.0 &
  43.6 &
  57.7 &
  65.4 &
  58.8 \\ 
{\color{black}Internvideo2.5-8B  \cite{wang2022internvideo} }
&57.3
&
  85.5 &
  33.3 &
  47.6 &
  56.4 &
  55.8 &
  63.5 &
  64.7 \\ 
  \midrule
   \rowcolor{gray!20}
\multicolumn{9}{c}{\textbf{\textit{RL-based Thinking MLLMs}}} \\ 
Video-R1-7B \cite{videor1} 
   &49.2
   &74.5
   &22.7
   &28.6
   &47.3
   &48.1
   &76.9
   &52.9
   \\ 
 \bottomrule
\end{tabular}
\caption{Performance of MLLMs on CVBench in cross-video complex reasoning tasks, evaluated across closed-source and open-source MLLMs. The tasks include: multi-view scene understanding (M. SU), multi-video temporal reasoning (M. TR), joint-video spatial navigation (J. SN), video difference captioning (VDC), cross-video causal reasoning (C. CR), joint-video summarization (J. S), and cross-video procedural transfer (C. PT).}
\label{Table1-3}
\end{table*}

\section{Benchmarked Results} 
\subsection{Experimental Settings}
\paragraph{Models.}  
Our evaluation comprehensively assesses various MLLMs, including closed-source and open-source models. A prominent model in our suite is Gemini~\cite{team2023gemini}, a closed-source model known for its sophisticated multimodal processing capabilities. On the open-source side, Qwen2.5-VL~\cite{videor1} stands out, offering transparency with its accessible codebase for further exploration and customization. We also evaluate advanced multimodal reasoning models such as Video-R1 \cite{videor1}, which leverages reinforcement learning to integrate multimodal information and perform complex cross-video reasoning tasks. For ablation study, we choose Qwen2.5VL-7B as the default model.

\paragraph{Settings.}
Unless otherwise specified, all open-source models are evaluated using 8 frames per video (e.g., totally 32 frames for a four-video QA example). For closed-source APIs, we use the default frame sampling rate. To enable multi-video inputs for mainstream models primarily trained on single-video tasks, we insert prompt tokens (i.e., ``{The video [video\_index].}”) before the visual tokens of each video to indicate their order and identifiers. The default resolution is set to 448×448 for all open-source models that support dynamic resolution; for models without such support, we use their default resolution as specified in this paper.

\begin{table*}[t!]
\centering
\begin{minipage}[t]{0.49\textwidth}
\scriptsize
\centering
\renewcommand\arraystretch{1.0}
\setlength{\tabcolsep}{0.55 mm}{
\begin{tabular}{lccccc}
\toprule
Categories 
& Average &
C. OR &
M. AR &
J. C &
C. EM \\
(QA pair numbers) 
&(249) &
(68) &
(46) &
(60)&
(75)\\ 
\midrule

\multicolumn{1}{l}{Human} 
&88.9
&85.7
&83.3
&80.0
& 100.0
\\ 
\multicolumn{1}{l}{Random Choice} 
&27.4
&25.0
&32.6
&25.0
&28.4
\\ 
\midrule
\rowcolor{gray!20}
\multicolumn{6}{c}{\textbf{\textit{Closed-source MLLMs}}} \\
\multicolumn{1}{l}{GPT-4o-mini}
&56.5
&58.8
& 60.9
& 36.7
& 58.1
\\
\multicolumn{1}{l}{GPT-4o}
&66.9
&66.2
& 76.1
& 51.7
& 66.2
\\
\multicolumn{1}{l}{Gemini-1.5-flash}
& 60.5
&63.2
& 69.6
& 35.0
& 60.8
\\
\multicolumn{1}{l}{Gemini-2.0-flash}
&64.5
& 67.6
& 73.9
& 38.3
& 63.5
\\ 
\midrule
\rowcolor{gray!20}
\multicolumn{6}{c}{\textbf{\textit{Open-source MLLMs}}} \\
Qwen2.5-Omni-7B \cite{xu2025qwen25omnitechnicalreport} &
54.4 &
52.9 &
60.9 &
35.0 &
59.5 \\
Qwen2.5-VL-7B \cite{bai2025qwen25vltechnicalreport} &
53.2 &
52.9 &
63.0 &
33.3 &
51.4 \\
LLaVA-OneVision-7B \cite{li2024llama} &
48.0 &
39.7 &
65.2 &
35.0 &
43.2 \\
VideoLLaMA3-7B \cite{cheng2024videollama} &
53.6 &
45.6 &
73.9 &
35.0 &
50.0 \\
InternVL2.5-8B \cite{chen2024internvl} &
60.9 &
58.8 &
76.1 &
43.3 &
54.1 \\
Phi-4-Multimodal-5B \cite{abouelenin2025phi} &
48.8 &
55.9 &
56.5 &
30.0 &
40.5 \\
Internvideo2.5-8B \cite{wang2022internvideo} &
59.3 &
60.3 &
67.4 &
43.3 &
56.8\\ 
\midrule
\rowcolor{gray!20}
\multicolumn{6}{c}{\textbf{\textit{RL-based Thinking MLLMs}}} \\
Video-R1-7B \cite{videor1} &54.4
&54.4
&63.0
&30.0
&40.5
\\ 
\bottomrule
\end{tabular}
}
\caption{Performance of MLLMs on CVBench regarding cross-video object association, evaluated across closed-source and open-source MLLMs. Tasks include: cross-video object recognition (C.OR), multi-video attribute recognition (M.AR), joint-video counting (J.C), and cross-video entity match~(C.EM). For human evaluation, we employed five annotators and reported the average accuracy. 
}
\label{Table1-1}
\end{minipage}
\hfill
\begin{minipage}[t]{0.49\textwidth}
\centering 
\scriptsize 
\renewcommand\arraystretch{1.01}
\setlength{\tabcolsep}{0.55 mm}{
\begin{tabular}{lccccc}
\toprule
Categories & Average &
C. AD &
C. SR &
M. KAR &
C. ER \\
(QA pair numbers) &
(369) &
(83) &
(149)&
(88)&
(49)
\\ 
\midrule
\multicolumn{1}{l}{Human} &92.7
 &100.0
&91.7
&83.3
&83.3

\\
\multicolumn{1}{l}{Random Choice}&33.8
&25.0
&53.0
&14.8
&24.5

\\
\midrule
\rowcolor{gray!20}
\multicolumn{6}{c}{\textbf{\textit{Closed-source MLLMs}}} \\
\multicolumn{1}{l}{GPT-4o-mini}
&60.0
& 56.0
& 60.4
&58.0
&69.4
\\
\multicolumn{1}{l}{GPT-4o}
&70.8
& 66.7
& 74.5
&65.9
&75.5
\\
\multicolumn{1}{l}{Gemini-1.5-flash}
& 68.7
& 64.3
&70.5
& 64.8
&77.6
\\
\multicolumn{1}{l}{Gemini-2.0-flash}
&67.0
& 65.5
& 67.8
& 67.0
& 67.3
\\ 
\midrule
\rowcolor{gray!20}
\multicolumn{6}{c}{\textbf{\textit{Open-source MLLMs}}} \\
Qwen2.5-Omni-7B \cite{xu2025qwen25omnitechnicalreport} &
58.9 &
53.6 &
61.1 &
58.0 &
63.3 \\
Qwen2.5-VL-7B \cite{bai2025qwen25vltechnicalreport} &
54.0 &
53.6 &
49.7 &
58.0 &
61.2 \\
LLaVA-OneVision-7B \cite{li2024llama} &
51.6 &
33.3 &
65.8 &
54.5 &
34.7 \\
VideoLLaMA3-7B \cite{cheng2024videollama} &
60.3 &
56.0 &
61.1 &
63.6 &
59.2 \\
InternVL2.5-8B \cite{chen2024internvl} &
62.4 &
57.1&
61.7 &
64.8 &
69.4 \\
Phi-4-Multimodal-5B \cite{abouelenin2025phi} &
51.9 &
44.0 &
54.4 &
59.1 &
44.9 \\
Internvideo2.5-8B \cite{wang2022internvideo} &
64.0 &
60.7 &
62.4 &
68.2 &
67.3\\ 
\midrule
\rowcolor{gray!20}
\multicolumn{6}{c}{\textbf{\textit{RL-based Thinking MLLMs}}} \\
Video-R1-7B \cite{videor1} &61.6
&59.5
&65.8
&55.7
&63.3
\\ 
\bottomrule
\end{tabular}}
\caption{Performance of MLLMs on CVBench regarding cross-video event association, evaluated across closed-source and open-source MLLMs. Tasks include: cross-video anomaly detection (C.AD), cross-video scene recognition (C.SR), multi-video key-action recognition (M.KAR), and cross-video event retrieval (C.ER). For human evaluation, we employed five annotators and reported the average accuracy. 
}
\label{Table1-2}
\end{minipage}
\end{table*}

\subsection{Results and Findings}

\paragraph{Overall performance.}

As presented in Table \ref{Table1-3}, Table \ref{Table1-1} and Table \ref{Table1-2}, our evaluation on CVBench reveals a significant performance gap between current MLLMs and human-level proficiency in cross-video reasoning. While human evaluators achieve an average accuracy of 91.3\%, the best-performing model, Gemini-2.0-flash, reaches only 69.4\%, leaving a substantial 21.9 point deficit. The results highlight a clear hierarchy of task difficulty. Models demonstrate pronounced weaknesses in tasks demanding precise temporal and causal alignment across videos, such as multi-video temporal reasoning (M. TR) and joint-video spatial navigation (J. SN), where even top models fail to surpass 60\% accuracy. In contrast, they exhibit relative competence in tasks that rely on aggregating high-level semantic information, like multi-view scene understanding (M. SU) and joint-video summarization (J. S), with several models achieving over 80\% accuracy.

\paragraph{Closed-source vs. open-source models.}
Closed-source models consistently outperform open-source models. This suggests that proprietary architectures may utilize more refined optimization techniques or superior capabilities in handling complex cross-video relationships, revealing a substantial gap between commercial-grade and publicly available MLLMs in this domain.

\begin{wraptable}{r}{7cm}
  \centering
  \vspace{0.5em}
  \begin{tabular}{lcccc}
    \toprule
    Run ID & \#1 & \#2 & \#3 & Avg. \\ 
    \midrule
    \multicolumn{5}{c}{\textbf{\textit{Disordered Input}}} \\ 
    Normal     & 46.8   & 46.8   & 46.8 & 46.8  \\
    Disordered & 43.0 	&41.3 	&42.7 	&42.3   \\
    $\Delta$     & 3.8 $\downarrow$ 	&5.5 $\downarrow$ 	&4.1 $\downarrow$ 	&4.5 $\downarrow$   \\ 
    \midrule
    \multicolumn{5}{c}{\textbf{\textit{Random Single Input}}} \\
    Normal     & 46.8  	&46.8  	&46.8  	&46.8    \\
    Single     & 37.2 	&37.5 	&37.8 	&37.5   \\
    $\Delta$     & 9.6 $\downarrow$ 	&9.3 $\downarrow$ 	&9.0 $\downarrow$ 	&9.3 $\downarrow$  \\ 
    \bottomrule
  \end{tabular}
  \caption{Ablation on input perturbations.}
  \label{Table2}
\end{wraptable}

\paragraph{Sensitivity to input perturbations.} 

For multi-video understanding, the way videos are presented may impact model performance. Table \ref{Table2} illustrates the performance differences when input videos are presented in regular order, disordered, or randomly selected. In the disordered input scenario, where the video order is randomized, the accuracy drops from 46.8\% to 42.3\%, with an average change of 4.5\%. For random single inputs, when models are presented with only a single video, performance decreases more sharply, from 46.8\% to 37.5\%, with an average change of 9.3\%. These findings highlight the sensitivity of models to input order, which further implies our benchmark focuses on fine-grained, temporal-aware relations between videos instead of merely global coarse perception.

\begin{wraptable}{r}{7cm}
    \centering
    \vspace{0.5em}
    \renewcommand\arraystretch{1.2}
    \setlength{\tabcolsep}{1.5 mm}{
    \begin{tabular}{lccc}
    \toprule
    \multicolumn{1}{l}{Resolution} & 224 $\times$ 224   & 448 $\times$ 448   & $\Delta$ \\ 
    \midrule
    \multicolumn{1}{l}{8f}               & 46.0      & 46.8      & 0.8 $\uparrow$  \\ 
    \multicolumn{1}{l}{16f}              & 47.3      & 47.6      & 0.3 $\uparrow$  \\ 
    \multicolumn{1}{l}{32f}              & 49.4      & 51.0      & 1.6 $\uparrow$  \\ 
    \multicolumn{1}{l}{$\Delta$}       & 1.7 $\uparrow$      & 2.1 $\uparrow$     & -   \\ 
    \bottomrule
    \end{tabular}
    \caption{Impact of frame number \& resolution.}
    \label{tab:table3}
    }
    
\end{wraptable}

\paragraph{Frame resolution.}
We investigate the impact of frame resolution using Qwen2.5-VL, considering different resolutions  and varying numbers of input frames, as shown in Table~\ref{tab:table3},. With 8 frames, accuracy increases slightly from 46.0\% at 224×224 to 46.8\% at 448×448 (+0.8\%), indicating that higher resolution leads to better performance. Similar trends are observed with 16 and 32 frames. Additionally, increasing the number of input frames consistently improves performance across both resolutions.

\begin{wrapfigure}{r}{1.05\linewidth}
    \centering
    \vspace{-1.5em}
    \includegraphics[width=1\linewidth]{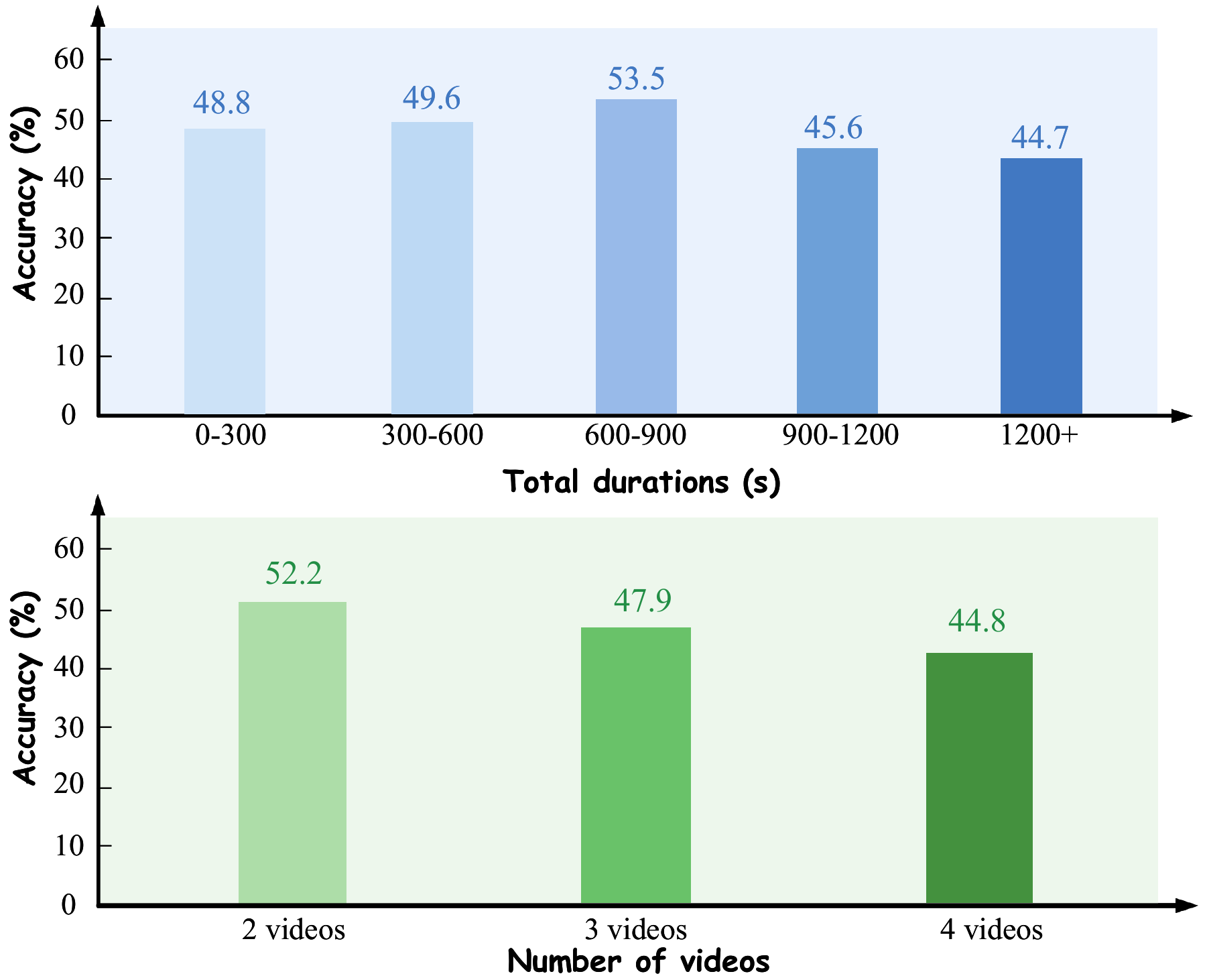}
    \caption{Impact of video number and duration.}
    \label{durationACC}
    \vspace{-1.5em}
\end{wrapfigure}

\paragraph{Number of video instances \& duration.}

Fig.~\ref{durationACC} illustrates how video duration and the number of videos affect model performance (Qwen2.5-VL). Accuracy improves with video length up to 900 seconds (48.8\% for 0–300s), but declines for longer videos (45.6\% for 900–1200s), likely due to increased sequence complexity and context management challenges.
We also highlight the difficulty of multi-video reasoning. Models that perform well on single-video tasks struggle with cross-video tasks. Shown in Fig.~\ref{durationACC}, the model achieves 52.2\%/47.9\%/44.8\% accuracy with two/three/four videos, showing increased complexity of multi-video tasks.

\begin{figure*}[t!]
    \includegraphics[width=0.99\linewidth]{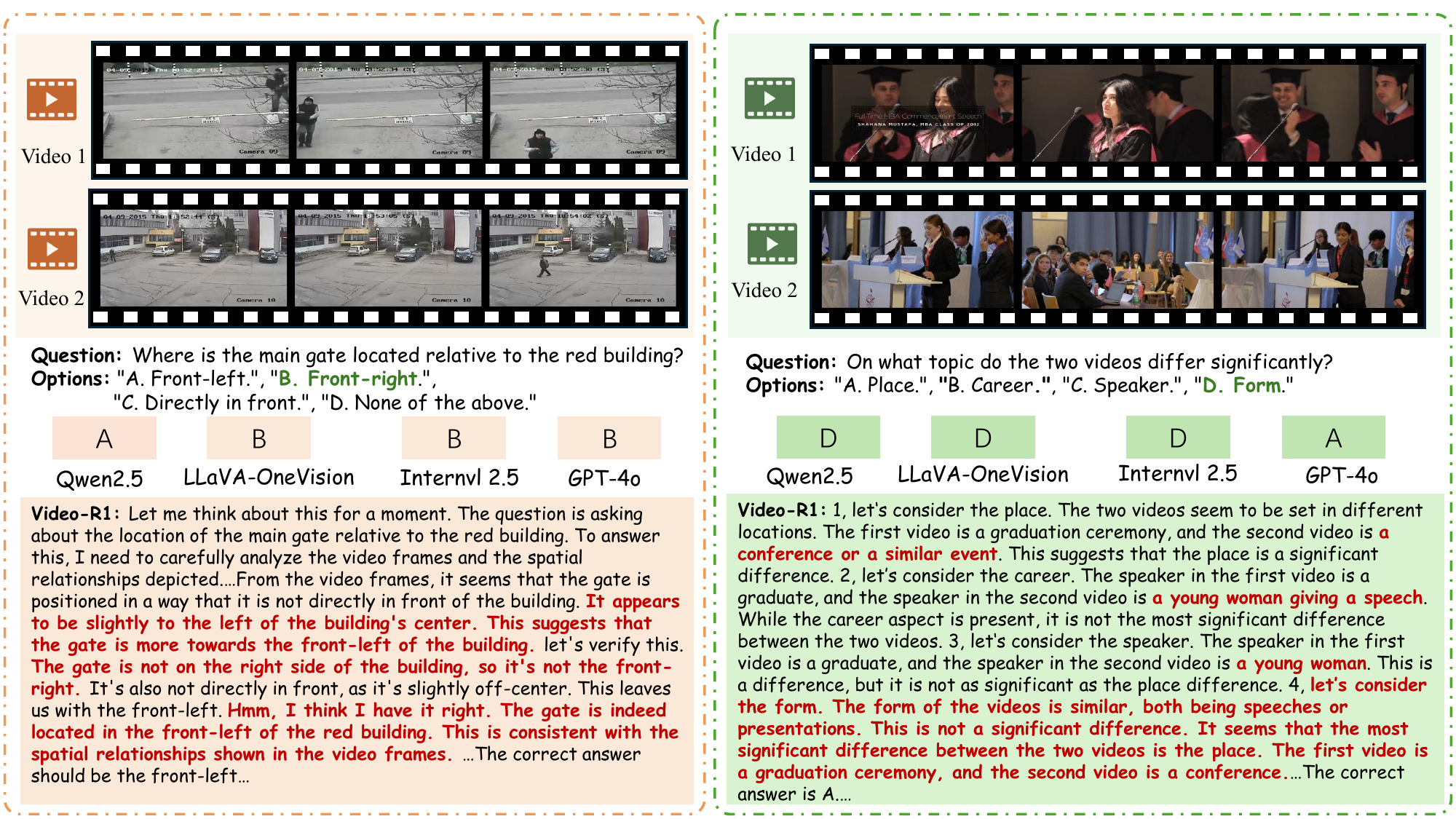}
    \caption{Qualitative cases of multi-video reasoning on CVBench based on Video-R1.}
    \label{Thinking}
\vspace{-1em}
\end{figure*}

\paragraph{Qualitative analysis of reasoning-based MLLMs.}
Qualitative analysis is crucial for diagnosing nuanced failures in multimodal large language models (MLLMs) beyond quantitative metrics. We examine the reasoning processes of Video-R1—a reinforcement learning-based MLLM—using the CVBench framework to uncover behavioral patterns revealing architectural limitations in cross-video reasoning. Empirical observations from Fig.~\ref{Thinking} illustrate critical performance aspects including hallucination and cognitive depth deficits. These insights complement CVBench’s quantitative results, where Video-R1 achieves only 49.2\% accuracy on complex reasoning tasks versus 91.3\% human performance, highlighting significant real-world applicability gaps.

The left sub-figure in Fig.~\ref{Thinking} demonstrates logical hallucination, where Video-R1 generates semantically coherent but logically inconsistent inferences with the multi-video context. For example, in a cross-video sports event association task, the model incorrectly assigns causal relationships between unrelated events (e.g., claiming an action in one video directly influences an outcome in another) despite absent spatiotemporal evidence. This failure stems from two core limitations: (1) inadequate cross-video attention mechanisms that disrupt entity state consistency across video boundaries, and (2) textual bias during inference, prioritizing linguistic priors over visual evidence. Such hallucinations intensify in tasks requiring precise spatiotemporal alignment (e.g., cross-video object tracking).

Conversely, the right sub-figure reveals cognitive depth deficits. Video-R1 struggles with hierarchical reasoning tasks demanding integration of commonsense knowledge and multi-view context. For instance, when synthesizing cooking procedures from multiple demonstration videos, the model identifies individual steps but fails to construct temporally coherent workflows. This aligns with CVBench’s finding that MLLMs exhibit surface-level understanding bias—excelling in object recognition but faltering in high-level reasoning (e.g., causal inference). These limitations amplify in asynchronous multi-camera scenarios where entities must be tracked across divergent viewpoints and timescales.

Current MLLMs like Video-R1 optimize for single-video processing, lacking modules for cross-video context retention or inter-video entity disambiguation. This gap impedes real-world deployment in dynamic, multi-source video reasoning applications.
\section{Conclusion}

This paper has introduced CVBench, a comprehensive diagnostic benchmark specifically designed to address the fundamental challenge of spatiotemporal relational pattern reasoning across multiple videos in multimodal large language models. By establishing a hierarchical taxonomy encompassing object association, event association, and complex reasoning tasks, CVBench provides a systematic framework for evaluating how models integrate and analyze relational patterns across diverse spatiotemporal contexts. Our extensive empirical assessment demonstrates that even state-of-the-art MLLMs exhibit significant limitations in tasks requiring consistent entity tracking, causal inference, and cross-video pattern integration, revealing critical gaps in current architectural approaches to multi-stream understanding.

The primary contribution of this work lies in the development of a standardized diagnostic framework that enables precise identification of model weaknesses, particularly in inter-video context retention and entity disambiguation within complex pattern spaces. Through its carefully curated dataset and rigorous evaluation protocol, CVBench establishes an essential tool for the pattern recognition community to benchmark progress and guide the development of next-generation architectures capable of authentic multi-stream relational reasoning. These findings underscore the necessity of developing specialized mechanisms that extend beyond single-video processing to address the unique challenges of cross-video pattern analysis.

While CVBench represents a substantial advancement in evaluating video-based multimodal reasoning, certain limitations related to task complexity and scope indicate important directions for future research. Primary efforts should focus on creating innovative cross-video attention mechanisms and memory-augmented architectures that can effectively model long-range spatiotemporal dependencies. Additionally, expanding the benchmark to incorporate more diverse real-world scenarios, including extreme conditions and domain-specific applications, will further advance the field of video pattern recognition.

\printbibliography

@article{ZHONG2025111035,
title = {A benchmark dataset and semantics-guided detection network for spatial–temporal human actions in urban driving scenes},
journal = {Pattern Recognition},
volume = {158},
pages = {111035},
year = {2025},
issn = {0031-3203}}

@article{SHAO2026112724,
title = {NEPose: A novel benchmark dataset with an improved framework for vision-based nasal endoscope pose estimation},
journal = {Pattern Recognition},
volume = {172},
pages = {112724},
year = {2026},
issn = {0031-3203}}

@article{ZHANG2026111925,
title = {Video and noise collaboratively guided semi-supervised diffusion model for video action detection},
journal = {Pattern Recognition},
volume = {169},
pages = {111925},
year = {2026},
issn = {0031-3203}}

@INPROCEEDINGS{Ventura11092338,
  author={Ventura, Lucas and Yang, Antoine and Schmid, Cordelia and Varol, Gül},
  booktitle={2025 IEEE/CVF Conference on Computer Vision and Pattern Recognition (CVPR)}, 
  title={Chapter-Llama: Efficient Chaptering in Hour-Long Videos with LLMs}, 
  year={2025},
  volume={},
  number={},
  pages={18947-18958}}

@ARTICLE{Ning11284911,
  author={Ning, Munan and Zhu, Bin and Xie, Yujia and Lin, Bin and Cui, Jiaxi and Yuan, Lu and Chen, Dongdong and Yuan, Li},
  journal={Computational Visual Media}, 
  title={Video-Bench: A comprehensive benchmark and toolkit for evaluating video-based large language models}, 
  year={2025},
  volume={},
  number={},
  pages={1-14}}

@INPROCEEDINGS{Wang11094563,
  author={Wang, Ziyang and Yu, Shoubin and Stengel-Eskin, Elias and Yoon, Jaehong and Cheng, Feng and Bertasius, Gedas and Bansal, Mohit},
  booktitle={2025 IEEE/CVF Conference on Computer Vision and Pattern Recognition (CVPR)}, 
  title={VideoTree: Adaptive Tree-based Video Representation for LLM Reasoning on Long Videos}, 
  year={2025},
  volume={},
  number={},
  pages={3272-3282}}

@INPROCEEDINGS{Yue11093104,
  author={Yue, Zhengrong and Zhuang, Shaobin and Li, Kunchang and Ding, Yanbo and Wang, Yali},
  booktitle={2025 IEEE/CVF Conference on Computer Vision and Pattern Recognition (CVPR)}, 
  title={V-Stylist: Video Stylization via Collaboration and Reflection of MLLM Agents}, 
  year={2025},
  volume={},
  number={},
  pages={3195-3205}}

@INPROCEEDINGS{Liu11094049,
  author={Liu, Zichen and Xu, Kunlun and Su, Bing and Zou, Xu and Peng, Yuxin and Zhou, Jiahuan},
  booktitle={2025 IEEE/CVF Conference on Computer Vision and Pattern Recognition (CVPR)}, 
  title={STOP: Integrated Spatial-Temporal Dynamic Prompting for Video Understanding}, 
  year={2025},
  volume={},
  number={},
  pages={13776-13786}}

@INPROCEEDINGS{Zhang11094660,
  author={Zhang, Huaxin and Xu, Xiaohao and Wang, Xiang and Zuo, Jialong and Huang, Xiaonan and Gao, Changxin and Zhang, Shanjun and Yu, Li and Sang, Nong},
  booktitle={2025 IEEE/CVF Conference on Computer Vision and Pattern Recognition (CVPR)}, 
  title={Holmes-VAU: Towards Long-term Video Anomaly Understanding at Any Granularity}, 
  year={2025},
  volume={},
  number={},
  pages={13843-13853}}

@INPROCEEDINGS{Ren10656135,
  author={Ren, Shuhuai and Yao, Linli and Li, Shicheng and Sun, Xu and Hou, Lu},
  booktitle={2024 IEEE/CVF Conference on Computer Vision and Pattern Recognition (CVPR)}, 
  title={TimeChat: A Time-sensitive Multimodal Large Language Model for Long Video Understanding}, 
  year={2024},
  volume={},
  number={},
  pages={14313-14323}}

@INPROCEEDINGS{Song10657734,
  author={Song, Enxin and Chai, Wenhao and Wang, Guanhong and Zhang, Yucheng and Zhou, Haoyang and Wu, Feiyang and Chi, Haozhe and Guo, Xun and Ye, Tian and Zhang, Yanting and Lu, Yan and Hwang, Jenq-Neng and Wang, Gaoang},
  booktitle={2024 IEEE/CVF Conference on Computer Vision and Pattern Recognition (CVPR)}, 
  title={MovieChat: From Dense Token to Sparse Memory for Long Video Understanding}, 
  year={2024},
  volume={},
  number={},
  pages={18221-18232}}

@INPROCEEDINGS{Fu11093290,
  author={Fu, Chaoyou and Dai, Yuhan and Luo, Yongdong and Li, Lei and Ren, Shuhuai and Zhang, Renrui and Wang, Zihan and Zhou, Chenyu and Shen, Yunhang and Zhang, Mengdan and Chen, Peixian and Li, Yanwei and Lin, Shaohui and Zhao, Sirui and Li, Ke and Xu, Tong and Zheng, Xiawu and Chen, Enhong and Shan, Caifeng and He, Ran and Sun, Xing},
  booktitle={2025 IEEE/CVF Conference on Computer Vision and Pattern Recognition (CVPR)}, 
  title={Video-MME: The First-Ever Comprehensive Evaluation Benchmark of Multi-modal LLMs in Video Analysis}, 
  year={2025},
  volume={},
  number={},
  pages={24108-24118}}

@article{ActivityNet2019,
  title={ActivityNet-QA: A Dataset for Understanding Complex Web Videos via Question Answering},
  author={Yu, Zhou and Xu, Dejing and Yu, Jun and Ting, Yu and Zhao, Zhou and Zhuang, Yueting and Tao, Dacheng},
  journal={Computer Vision and Pattern Recognition},
  year={2019}
}

@article{team2023gemini,
  title={Gemini: a family of highly capable multimodal models},
  author={Team, Gemini and Anil, Rohan and Borgeaud, Sebastian and Alayrac, Jean-Baptiste and Yu, Jiahui and Soricut, Radu and Schalkwyk, Johan and Dai, Andrew M and Hauth, Anja and Millican, Katie and others},
  journal={arXiv preprint arXiv:2312.11805},
  year={2023}
}

@article{fu2024video,
  title={Video-mme: The first-ever comprehensive evaluation benchmark of multi-modal llms in video analysis},
  author={Fu, Chaoyou and Dai, Yuhan and Luo, Yongdong and Li, Lei and Ren, Shuhuai and Zhang, Renrui and Wang, Zihan and Zhou, Chenyu and Shen, Yunhang and Zhang, Mengdan and others},
  journal={arXiv preprint arXiv:2405.21075},
  year={2024}
}

@article{videor1,
  title={Video-R1: Reinforcing Video Reasoning in MLLMs},
  author={Feng, Kaituo and Gong, Kaixiong and Li, Bohao and Guo, Zonghao and Wang, Yibing and Peng, Tianshuo and Wang, Benyou and Yue, Xiangyu},
  journal={arXiv preprint arXiv:2503.21776},
  year={2025}
}

@inproceedings{li2024llama,
  title={Llama-vid: An image is worth 2 tokens in large language models},
  author={Li, Yanwei and Wang, Chengyao and Jia, Jiaya},
  booktitle={European Conference on Computer Vision},
  pages={323--340},
  year={2024},
  organization={Springer}
}

@article{cheng2024videollama,
  title={Videollama 2: Advancing spatial-temporal modeling and audio understanding in video-llms},
  author={Cheng, Zesen and Leng, Sicong and Zhang, Hang and Xin, Yifei and Li, Xin and Chen, Guanzheng and Zhu, Yongxin and Zhang, Wenqi and Luo, Ziyang and Zhao, Deli and others},
  journal={arXiv preprint arXiv:2406.07476},
  year={2024}
}

@article{tempcompass2024,
  title={Tempcompass: Do video llms really understand videos?},
  author={Liu, Yuanxin and Li, Shicheng and Liu, Yi and Wang, Yuxiang and Ren, Shuhuai and Li, Lei and Chen, Sishuo and Sun, Xu and Hou, Lu},
  journal={arXiv preprint arXiv:2403.00476},
  year={2024}
}

@article{egoschema2023,
  title={Egoschema: A diagnostic benchmark for very long-form video language understanding},
  author={Mangalam, Karttikeya and Akshulakov, Raiymbek and Malik, Jitendra},
  journal={Advances in Neural Information Processing Systems},
  volume={36},
  pages={46212--46244},
  year={2023}
}

@inproceedings{xu2017video,
  title={Video question answering via gradually refined attention over appearance and motion},
  author={Xu, Dejing and Zhao, Zhou and Xiao, Jun and Wu, Fei and Zhang, Hanwang and He, Xiangnan and Zhuang, Yueting},
  booktitle={Proceedings of the 25th ACM international conference on Multimedia},
  pages={1645--1653},
  year={2017}
}

@inproceedings{xiao2021next,
  title={Next-qa: Next phase of question-answering to explaining temporal actions},
  author={Xiao, Junbin and Shang, Xindi and Yao, Angela and Chua, Tat-Seng},
  booktitle={Proceedings of the IEEE/CVF conference on computer vision and pattern recognition},
  pages={9777--9786},
  year={2021}
}

@inproceedings{chen2024autoeval,
  title={Autoeval-video: An automatic benchmark for assessing large vision language models in open-ended video question answering},
  author={Chen, Xiuyuan and Lin, Yuan and Zhang, Yuchen and Huang, Weiran},
  booktitle={European Conference on Computer Vision},
  pages={179--195},
  year={2024},
  organization={Springer}
}

@ARTICLE{Wu2025,
  author={Wu, Wenhao and Wang, Xiaohan and Luo, Haipeng and Wang, Jingdong and Yang, Yi and Ouyang, Wanli},
  journal={IEEE Transactions on Pattern Analysis and Machine Intelligence}, 
  title={Cap4Video++: Enhancing Video Understanding With Auxiliary Captions}, 
  year={2025},
  volume={47},
  number={7},
  pages={5223-5237}}

@article{CAI2025111670,
title = {MLLM as video narrator: Mitigating modality imbalance in video moment retrieval},
author = {Weitong Cai and Jiabo Huang and Shaogang Gong and Hailin Jin and Yang Liu},
journal = {Pattern Recognition},
volume = {166},
pages = {111670},
year = {2025},
issn = {0031-3203}}

@ARTICLE{11008449,
  author={Ge, Qihang and Sun, Wei and Zhang, Yu and Li, Yunhao and Ji, Zhongpeng and Sun, Fengyu and Jui, Shangling and Min, Xiongkuo and Zhai, Guangtao},
  journal={IEEE Transactions on Circuits and Systems for Video Technology}, 
  title={LMM-VQA: Advancing Video Quality Assessment With Large Multimodal Models}, 
  year={2025},
  volume={35},
  number={11},
  pages={11083-11096}}

@article{ZHENG2025111319,
title = {Region-aware mutual relational knowledge distillation for semantic segmentation},
author = {Haowen Zheng and Xuxin Lin and Hailun Liang and Benjia Zhou and Yanyan Liang},
journal = {Pattern Recognition},
volume = {161},
pages = {111319},
year = {2025},
issn = {0031-3203}}

@INPROCEEDINGS{Li10658165,
  author={Li, Kunchang and Wang, Yali and He, Yinan and Li, Yizhuo and Wang, Yi and Liu, Yi and Wang, Zun and Xu, Jilan and Chen, Guo and Lou, Ping and Wang, Limin and Qiao, Yu},
  booktitle={2024 IEEE/CVF Conference on Computer Vision and Pattern Recognition (CVPR)}, 
  title={MVBench: A Comprehensive Multi-modal Video Understanding Benchmark}, 
  year={2024},
  volume={},
  number={},
  pages={22195-22206}}

@INPROCEEDINGS{11093290,
  author={Fu, Chaoyou and Dai, Yuhan and Luo, Yongdong and Li, Lei and Ren, Shuhuai and Zhang, Renrui and Wang, Zihan and Zhou, Chenyu and Shen, Yunhang and Zhang, Mengdan and Chen, Peixian and Li, Yanwei and Lin, Shaohui and Zhao, Sirui and Li, Ke and Xu, Tong and Zheng, Xiawu and Chen, Enhong and Shan, Caifeng and He, Ran and Sun, Xing},
  booktitle={2025 IEEE/CVF Conference on Computer Vision and Pattern Recognition (CVPR)}, 
  title={Video-MME: The First-Ever Comprehensive Evaluation Benchmark of Multi-modal LLMs in Video Analysis}, 
  year={2025},
  volume={},
  number={},
  pages={24108-24118}}

@ARTICLE{Yang11037495,
  author={Yang, Xuzheng and Liu, Junzhuo and Wang, Peng and Wang, Guoqing and Yang, Yang and Shen, Heng Tao},
  journal={IEEE Transactions on Pattern Analysis and Machine Intelligence}, 
  title={New Dataset and Methods for Fine-Grained Compositional Referring Expression Comprehension via Specialist-MLLM Collaboration}, 
  year={2025},
  volume={47},
  number={10},
  pages={8598-8612}}

@INPROCEEDINGS{11095110,
  author={Geng, Tiantian and Zhang, Jinrui and Wang, Qingni and Wang, Teng and Duan, Jinming and Zheng, Feng},
  booktitle={2025 IEEE/CVF Conference on Computer Vision and Pattern Recognition (CVPR)}, 
  title={LongVALE: Vision-Audio-Language-Event Benchmark Towards Time-Aware Omni-Modal Perception of Long Videos}, 
  year={2025},
  volume={},
  number={},
  pages={18959-18969}}

@ARTICLE{Zhang10547418,
  author={Zhang, Wei and Cai, Miaoxin and Zhang, Tong and Zhuang, Yin and Mao, Xuerui},
  journal={IEEE Transactions on Geoscience and Remote Sensing}, 
  title={EarthGPT: A Universal Multimodal Large Language Model for Multisensor Image Comprehension in Remote Sensing Domain}, 
  year={2024},
  volume={62},
  number={},
  pages={1-20}}

@article{wang2022internvideo,
  title={Internvideo: General video foundation models via generative and discriminative learning},
  author={Wang, Yi and Li, Kunchang and Li, Yizhuo and He, Yinan and Huang, Bingkun and Zhao, Zhiyu and Zhang, Hongjie and Xu, Jilan and Liu, Yi and Wang, Zun and others},
  journal={arXiv preprint arXiv:2212.03191},
  year={2022}
}

@article{abouelenin2025phi,
  title={Phi-4-mini technical report: Compact yet powerful multimodal language models via mixture-of-loras},
  author={Abouelenin, Abdelrahman and Ashfaq, Atabak and Atkinson, Adam and Awadalla, Hany and Bach, Nguyen and Bao, Jianmin and Benhaim, Alon and Cai, Martin and Chaudhary, Vishrav and Chen, Congcong and others},
  journal={arXiv preprint arXiv:2503.01743},
  year={2025}
}

@inproceedings{chen2024internvl,
  title={Internvl: Scaling up vision foundation models and aligning for generic visual-linguistic tasks},
  author={Chen, Zhe and Wu, Jiannan and Wang, Wenhai and Su, Weijie and Chen, Guo and Xing, Sen and Zhong, Muyan and Zhang, Qinglong and Zhu, Xizhou and Lu, Lewei and others},
  booktitle={Proceedings of the IEEE/CVF conference on computer vision and pattern recognition},
  pages={24185--24198},
  year={2024}
}

@misc{xu2025qwen25omnitechnicalreport,
      title={Qwen2.5-Omni Technical Report}, 
      author={Jin Xu and Zhifang Guo and Jinzheng He and Hangrui Hu and Ting He and Shuai Bai and Keqin Chen and Jialin Wang and Yang Fan and Kai Dang and Bin Zhang and Xiong Wang and Yunfei Chu and Junyang Lin},
      year={2025},
      eprint={2503.20215},
      archivePrefix={arXiv},
      primaryClass={cs.CL},
    }

@misc{bai2025qwen25vltechnicalreport,
      title={Qwen2.5-VL Technical Report}, 
      author={Shuai Bai and Keqin Chen and Xuejing Liu and Jialin Wang and Wenbin Ge and Sibo Song and Kai Dang and Peng Wang and Shijie Wang and Jun Tang and Humen Zhong and Yuanzhi Zhu and Mingkun Yang and Zhaohai Li and Jianqiang Wan and Pengfei Wang and Wei Ding and Zheren Fu and Yiheng Xu and Jiabo Ye and Xi Zhang and Tianbao Xie and Zesen Cheng and Hang Zhang and Zhibo Yang and Haiyang Xu and Junyang Lin},
      year={2025},
      eprint={2502.13923},
      archivePrefix={arXiv},
      primaryClass={cs.CV},
}
\clearpage
\appendix

\section{Dataset Collection}

\subsection{Task Definitions}
While single-video tasks assess a model's comprehension of individual videos, CVBench evaluates cross-video relational reasoning. This critical capability enables MLLMs to advance from controlled lab settings to robust real-world deployment.
CVBench introduces 15 carefully curated evaluation dimensions that collectively measure a model's capacity for inter-video relational reasoning, each targeting a unique but complementary aspect of cross-video understanding. An overview of these tasks, their concepts, differences, and example questions is provided in Table \ref{tab:task2}.


\begin{table}[htbp]
\centering
\begin{tcolorbox}[colback=gray!10, colframe=white]
\small
\textcolor{blue}{\textit{Task 1: Cross-video anomaly detection}} 

\textbf{Concept:} The comparative analysis of multi-source video data identifies anomalous segments deviating from normal patterns through global spatiotemporal feature and behavioral pattern comparisons.

\textbf{Difference:} Analyzes anomaly saliency through multi-video event richness and viewpoint coherence rather than single-video perspectives.

\textbf{Example Questions:}
\begin{itemize}[leftmargin=*, nosep]

    \item {\textit{Question}:} Which video features individuals directly engaged in construction work rather than showcasing materials or interior design?  
    
    {\textit{Option}:} A. Video 1, B. Video 2, C. Video 3, D. Video 4
    
\end{itemize}
\ \\


\textcolor{blue}{\textit{Task 2: Cross-video scene recognition}} 

\textbf{Concept:} Fusion of semantic features from multiple videos to locate scene segments matching target spatiotemporal attributes.

\textbf{Difference:} Focuses on cross-video differences rather than single-video similarities. 

\textbf{Example Questions:}
\begin{itemize}[leftmargin=*, nosep]
    \item {\textit{Question}:} ``Did both videos feature a team successfully scoring a decisive goal or point during the match's closing moments?'' 
    
    {\textit{Option}:} Yes/No.
\end{itemize}
 
\ \\

\textcolor{blue}{\textit{Task 3:  Multi-video key-action recognition}}

\textbf{Concept:} This category involves spatiotemporal alignment of motion trajectories across videos to identify differences in action execution, supporting action quality assessment and optimization.

\textbf{Difference:}  It emphasizes modeling the spatiotemporal consistency of actions across multiple videos, rather than classifying actions within a single video.
\textbf{Example Questions:}

\begin{itemize}[leftmargin=*, nosep]
    \item  {\textit{Question}:} What is the key action difference between the presenter in Video 1 and the presenter in Video 2?
    
    {\textit{Option}:}  A. In Video 1, holding products is key, while in Video 2, gesturing with hands dominates., B. In Video 1, gesturing is emphasized more, while in Video 2, interacting with products is dominant., C. In both, sitting is a primary action, but outdoor walking is unique to Video 2., D. Video 2 uses textual prompts for interactions, while Video 1 does not.
      
\end{itemize}



\end{tcolorbox}

\label{tab:task2}
\end{table}

\begin{table}[t!]
\centering

\begin{tcolorbox}[colback=gray!10, colframe=white, width=1\textwidth, boxrule=0pt, arc=0pt]
\small
\textcolor{blue}{\textit{Task 4: Cross-video event retrieval}}

\textbf{Concept:} This category involves rapidly locating segments across multiple videos that meet specific event elements, enabling effective filtering and selection.

\textbf{Difference:}  The events are distributed across different videos, rather than being contained within a single video.

\textbf{Example Questions:}
\begin{itemize}[leftmargin=*, nosep]
    \item  {\textit{Question}:} Which video shows the celebration of a team after successfully leading or winning in the match?\\
{\textit{Option}:}  A. Video 1.", "B. Video 2.", "C. Both of them.", "D. None of them.

\end{itemize}
 \ \\
 
\textcolor{blue}{\textit{Task 5: Cross-video object recognition}}

\textbf{Concept:} This category involves fusing multi-view object features to address challenges such as occlusion and deformation, enabling consistent identity recognition of objects across videos.

\textbf{Difference:}  It leverages multi-source information to compensate for the limitations of single-video perspectives.

\textbf{Example Questions:}
\begin{itemize}[leftmargin=*, nosep]
    \item  {\textit{Question}:} Which item is commonly recognized across both Videos 1 and 2 despite differences in design aesthetic?
  
    {\textit{Option}:} A. Bamboo plants.", "B. Distressed wooden door.", "C. Umbrella stands.", "D. Metalwork tools.

\end{itemize}
\ \\

\textcolor{blue}{\textit{Task 6: Multi-video attribute recognition}}

\textbf{Concept:} This category involves confirming and extracting the attributes (such as texture, color, function, relationship, characteristic and shape) of a specific target across multiple videos, capturing the same target in different states.

\textbf{Difference:} It focuses on cross-video attribute state transitions, rather than static attribute recognition within a single frame.

\textbf{Example Questions:}
\begin{itemize}[leftmargin=*, nosep]
    \item  {\textit{Question}:} What color predominantly features in the attire of winning players across both sporting events?

       {\textit{Option}:} "A. White.", "B. Red.", "C. Blue.", "D. Green."

\end{itemize}
\ \\

\textcolor{blue}{\textit{Task 7: Joint-video counting}}

\textbf{Concept:} This category involves the precise identification and statistical analysis of the same target across multiple videos.

\textbf{Difference:} The targets are distributed across multiple videos rather than a single video, requiring cross-video unified authentication of the targets, while eliminating limitations of single-video perspectives (such as occlusion). Targets may also be abstract, such as events.

\textbf{Example Questions:}
\begin{itemize}[leftmargin=*, nosep]
    \item  {\textit{Question}:} How many monitors other than cell phones are in all the videos?

      {\textit{Option}:}  A. 1, B. 2, C. 3, D. 4.
       
\end{itemize}

\ \\

\textcolor{blue}{\textit{Task 8: Cross-video entity matching}}

\textbf{Concept:} This category involves making similarity judgments of entities across multiple videos with varying spatiotemporal conditions (different space and time, same space and different times, different space and different times), such as identifying criminal suspects.

\end{tcolorbox}

\label{tab:task2}
\end{table}

\begin{table}[htbp]
\centering

\begin{tcolorbox}[colback=gray!10, colframe=white, width=0.99\textwidth, boxrule=0pt, arc=0pt]
\small

\textcolor{blue}{\textit{Task 10:  Multi-video temporal reasoning}}

\textbf{Concept:} This category involves integrating multiple videos and making judgments about hidden logical relationships at specific times, such as predicting the future or sorting events.

\textbf{Difference:} It addresses the asynchronous timestamp issue in single videos, enabling global temporal modeling.

\textbf{Example Questions:}
\begin{itemize}[leftmargin=*, nosep]
    \item  {\textit{Question}:} Based on the interconnections, what is the correct chronological order of the criminal case depicted across the videos?
        
    {\textit{Option}:}   A. 1-2-3-4.", "B. 4-1-2-3.", "C. 1-2-4-3.", "D. 3-1-4-2.
   
\end{itemize}

\ \\

\textcolor{blue}{\textit{Task 11: Joint-video spatial navigating}}

\textbf{Concept:} This category involves fusing multi-view geometric information to construct a 3D spatial semantic map, supporting cross-view path planning.

\textbf{Difference:} Spatial registration and joint reasoning of multi-source visual data.

\textbf{Example Questions:}
\begin{itemize}[leftmargin=*, nosep]
    \item  {\textit{Question}:} Where are the girl's two wardrobes located in the room across all videos?
   
 {\textit{Option}:}   A. Head to the top-right corner to enter the kitchen and stove is located in the top-right corner of the kitchen, B. Head to the top-left corner to enter the kitchen and stove is located in the top-left corner of the kitchen, C. Head to the 
\end{itemize}
\ \\

\textcolor{blue}{\textit{Task 12: Video difference caption}}

\textbf{Concept:} This category involves fine-grained cross-video comparison, identifying differences across multiple videos in dimensions such as event progression and object states.

\textbf{Difference:} It emphasizes the identification of differences in dynamic processes, rather than static image comparison.

\textbf{Example Questions:}
\begin{itemize}[leftmargin=*, nosep]
    \item  {\textit{Question}:} What differentiates Video 1's organization 
 method from Video 2 and Video 3?

     {\textit{Option}:} A. Uses a rotating tray., B. Combines different gadgets., C. Focuses on vertical space., D. Emphasizes utensil storage.
\end{itemize}

\ \\

\textcolor{blue}{\textit{Task 13:  Cross-video counterfactual reasoning}}

\textbf{Concept:} Based on the spatiotemporal factual foundation from multiple videos, this category constructs a causal inference chain for a virtual scenario.

\textbf{Difference:} Causal effect estimation based on multi-source observational data, overcoming the limitations of single-video causal inference.

\textbf{Example Questions:}
\begin{itemize}[leftmargin=*, nosep]
    \item  {\textit{Question}:}  What element remains unchanged across the videos, regardless of the setting?

    {\textit{Option}:}  A. Clothing style., B. Dance type., C. Footwear focus., D. Presence of music.

\end{itemize}
\ \\

\textcolor{blue}{\textit{Task 14: Cross-video object recognition}}

\textbf{Concept:} This category involves fusing multi-view object features to address challenges such as occlusion and deformation, enabling consistent identity recognition of objects across videos.

\textbf{Difference:}  It leverages multi-source information to compensate for the limitations of single-video perspectives.

\textbf{Example Questions:}
\begin{itemize}[leftmargin=*, nosep]
    \item  {\textit{Question}:} Which item is commonly recognized across both Videos 1 and 2 despite differences in design aesthetic?
  
    {\textit{Option}:} A. Bamboo plants.", "B. Distressed wooden door.", "C. Umbrella stands.", "D. Metalwork tools.

\end{itemize}
\ \\

\textcolor{blue}{\textit{Task 15: Cross-video event retrieval}}

\textbf{Concept:} This category involves rapidly locating segments across multiple videos that meet specific event elements, enabling effective filtering and selection.

\textbf{Difference:}  The events are distributed across different videos, rather than being contained within a single video.

\textbf{Example Questions:}
\begin{itemize}[leftmargin=*, nosep]
    \item  {\textit{Question}:} Which video shows the celebration of a team after successfully leading or winning in the match?\\
{\textit{Option}:}  A. Video 1.", "B. Video 2.", "C. Both of them.", "D. None of them.

\end{itemize}

\end{tcolorbox}
\caption{Overview of cross-video tasks, concepts, differences, and example questions.}
\label{tab:task2}
\end{table}

\subsection{QA Generation}
\begin{figure}
    \centering
    \includegraphics[width=1\linewidth]{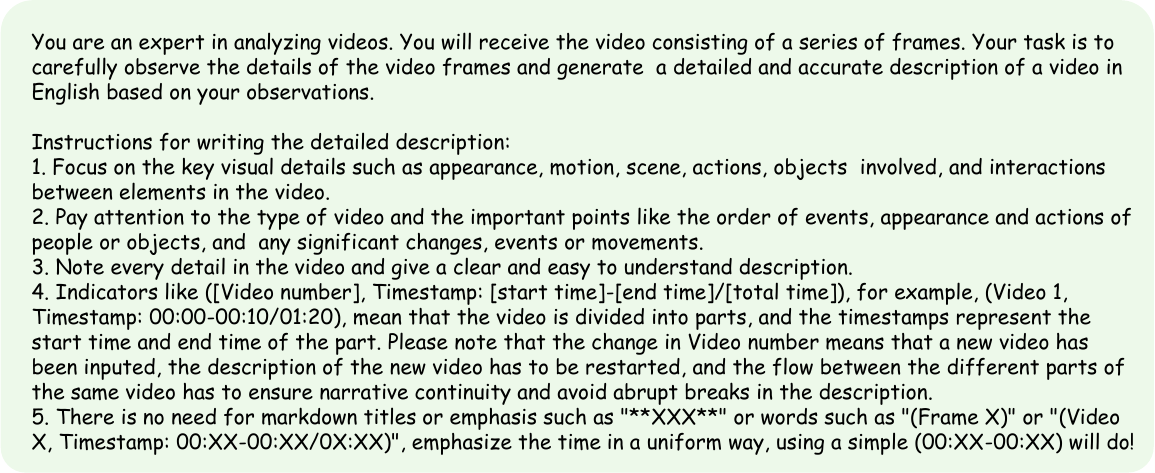}
    \caption{Prompt template for generating cohesive video summaries with GPT-4o.
generate a cohesive summary of the following video by merging the segment-wise captions into a comprehensive temporal description that fully encapsulates the video's content.}
    \label{Segprompt}
\end{figure}

\begin{figure}
    \centering
    \includegraphics[width=1\linewidth]{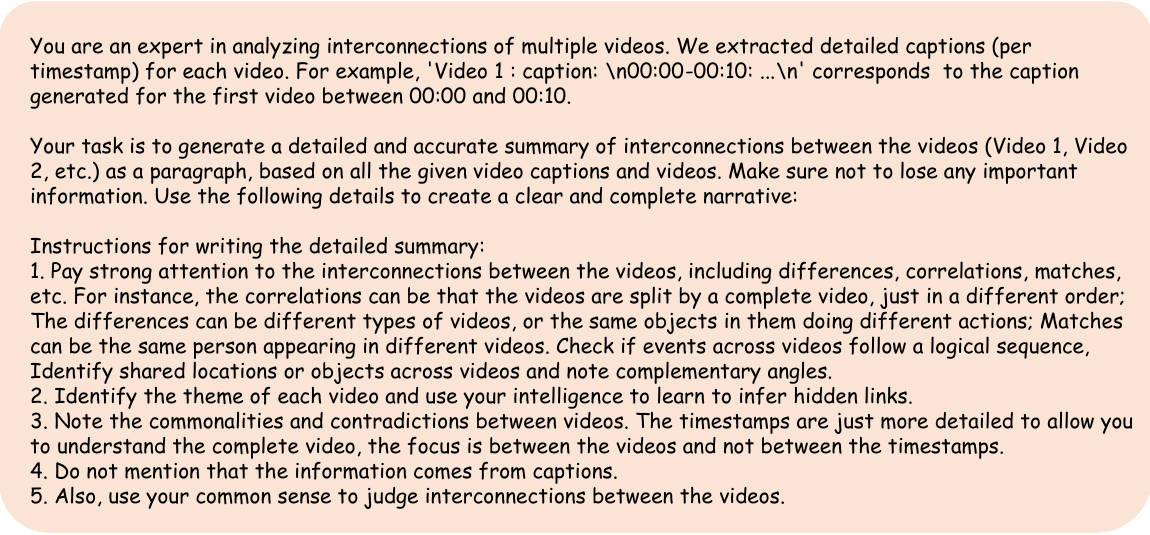}
    \caption{Cross-video relation extraction prompt. Analyze the interconnections between multiple videos by combining their captions. }
    \label{crossprompt}
\end{figure}

\begin{figure}
    \centering
    \includegraphics[width=1\linewidth]{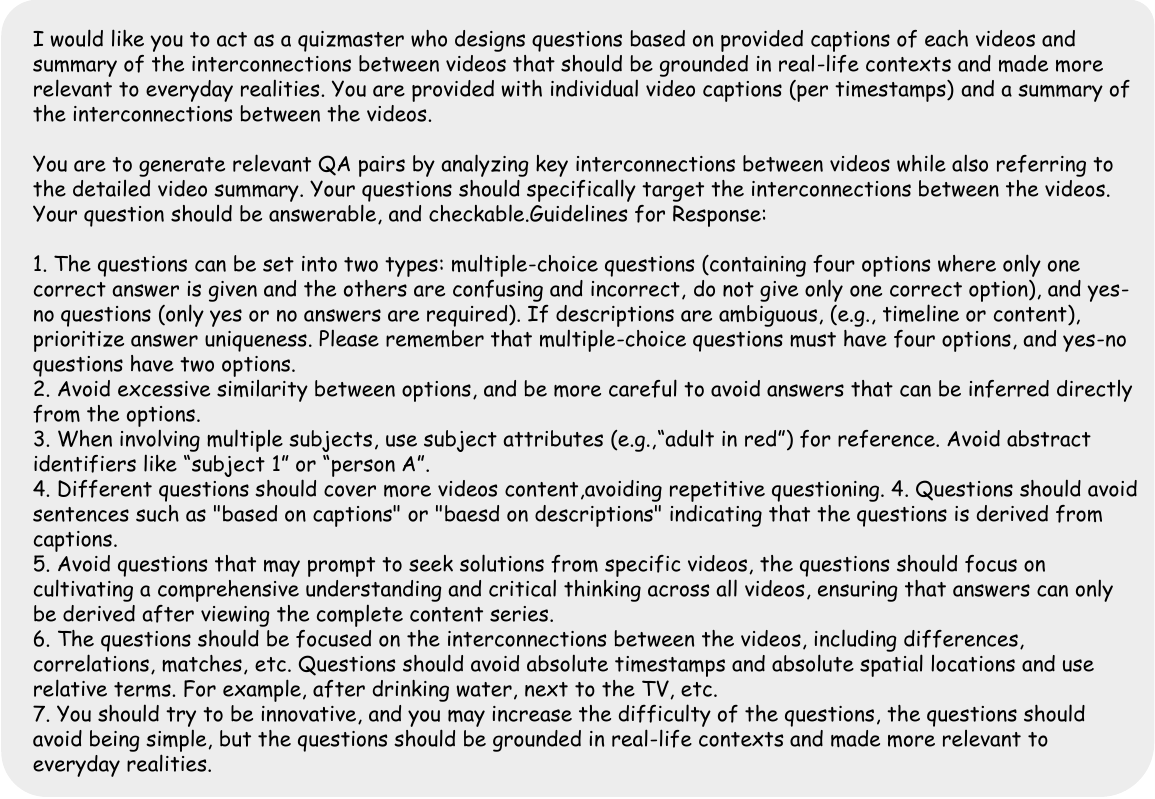}
    \caption{QA pair  prompt. Create quiz questions based on the interconnections between video captions.}
    \label{QAprompt}
\end{figure}

We propose a hierarchical framework for generating cross-video question-answer pairs with three key stages for comprehensive evaluation. In the first stage, adaptive video segmentation is combined with GPT-4o to generate fine-grained spatiotemporal annotations, capturing critical elements such as object interactions and action sequences. The prompt template for generating cohesive video summaries is shown in Figure~\ref{Segprompt}. In the second stage, a relationship extraction module constructs a cross-video association graph, identifying multidimensional relationships such as object evolution and event continuity. The cross-video relation extraction prompt is depicted in Figure~\ref{crossprompt}. Based on 15 professional templates, the system generates diverse questions, ranging from basic matching to advanced reasoning, ensuring quality through multi-model validation and difficulty calibration. The QA pair prompt template is shown in Figure~\ref{QAprompt}.

This framework integrates visual content analysis with logical reasoning, featuring a multi-layered quality control system that includes multi-model consensus validation and hallucination detection to ensure the verifiability and distinctiveness of the generated questions. Ultimately, the framework provides a comprehensive evaluation spectrum, ranging from single-hop to commonsense reasoning, with multidimensional metrics such as accuracy and reasoning depth. Figures~\ref{Segprompt}, \ref{crossprompt}, and \ref{QAprompt} present standardized prompt templates, ensuring the systematization and reproducibility of the evaluation process.

\subsection{More CVBench Examples Visualization}
\begin{figure}
    \centering
    \includegraphics[width=1\linewidth]{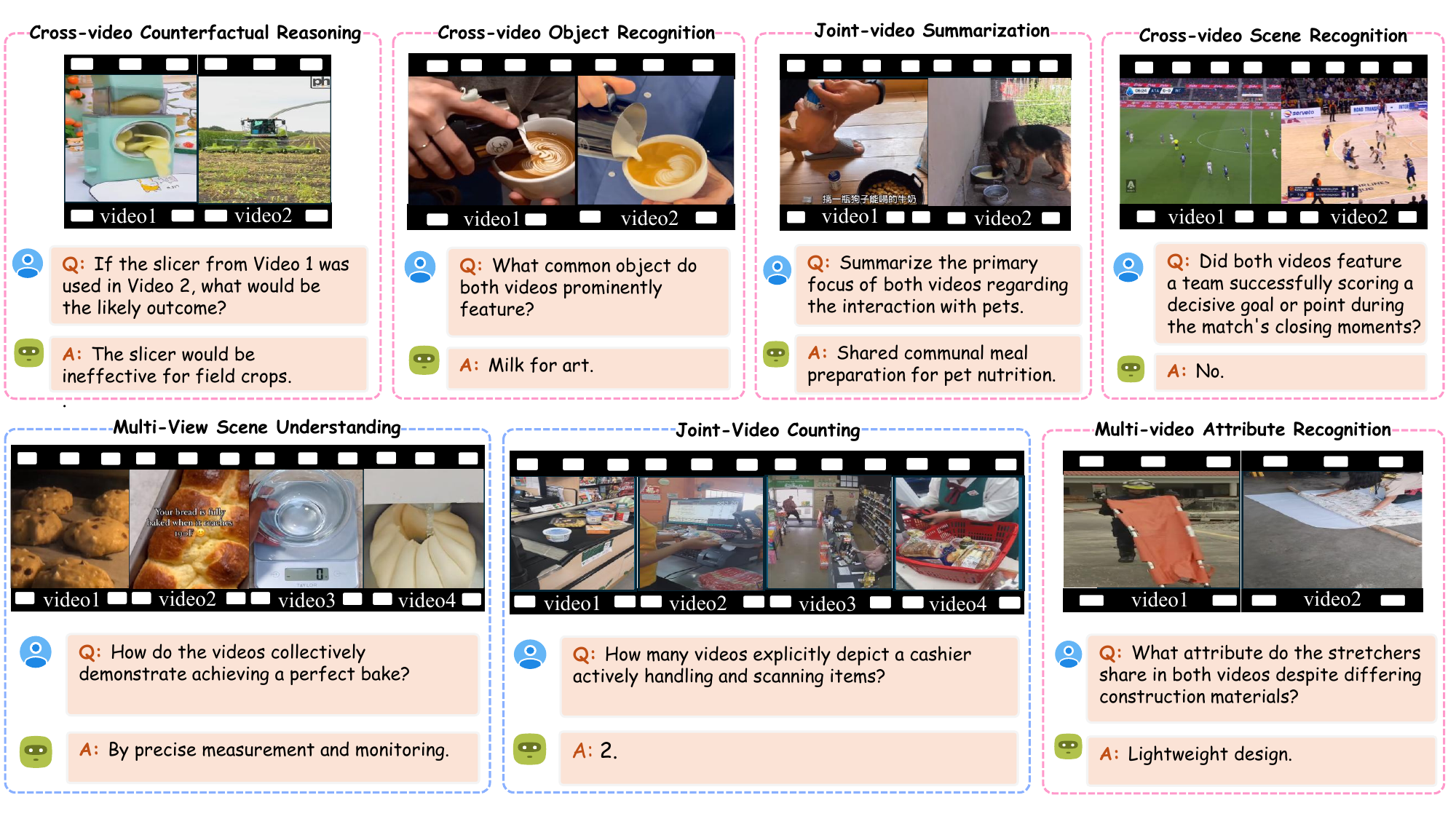}
    \caption{More examples of CVBench.}
    \label{Examples of CVBench}
\end{figure}

CVBench's 15 carefully designed subcategories of videos and questions evaluate key dimensions such as spatiotemporal correlation, entity-event association, and causal-logical reasoning. These tasks cover various challenges, from tracking object consistency across cameras to inferring causal relationships between multiple videos, detecting spatial annotation conflicts, and performing cross-modal data fusion. Each subcategory presents complex tasks that closely resemble real-world scenarios.
More examples of these tasks are provided in Figure \ref{Examples of CVBench}.

\end{document}